\documentclass[aps,prb,citeautoscript,twocolumn,showpacs,superscriptaddress]{revtex4-1}  % for publication

\usepackage{graphicx}  % needed for figures
\usepackage{dcolumn}   % needed for some tables
\usepackage{bm}        % for math
\usepackage{amssymb}   % for math
\usepackage{amsmath}   % for math
\usepackage{algorithm}
\usepackage{algpseudocode}
\usepackage[dvipsnames]{xcolor}
\usepackage{soul}

\newcommand{\revision}[1]{#1}

\begin{document}

\title{Building high accuracy emulators for scientific simulations\\with deep neural architecture search\\
%{\small(Short title: Accelerating simulations by two billion times)}
}

\author{M. F. Kasim}
\email{muhammad.kasim@physics.ox.ac.uk}
\affiliation{Clarendon Laboratory, Department of Physics, University of Oxford, Parks Road, Oxford, UK}
\author{D. Watson-Parris}
\author{L. Deaconu}
\affiliation{Atmospheric, Oceanic and Planetary Physics, Department of Physics, University of Oxford, Oxford, UK}
\author{S. Oliver}
\affiliation{Department of Earth Sciences, University of Oxford, Oxford, UK}
\author{P. Hatfield}
\affiliation{Clarendon Laboratory, Department of Physics, University of Oxford, Parks Road, Oxford, UK}

\author{D. H. Froula}
\affiliation{Laboratory for Laser Energetics, University of Rochester, New York, USA}

\author{G. Gregori}
\affiliation{Clarendon Laboratory, Department of Physics, University of Oxford, Parks Road, Oxford, UK}
\author{M. Jarvis}
\affiliation{Denys Wilkinson Building, Department of Physics, University of Oxford, Keble Road, Oxford, UK}
\author{S. Khatiwala}
\affiliation{Department of Earth Sciences, University of Oxford, Oxford, UK}
\author{J. Korenaga}
\affiliation{Department of Geology and Geophysics, Yale University, New Haven, Connecticut, USA}
\author{J. Topp-Mugglestone}
\affiliation{Clarendon Laboratory, Department of Physics, University of Oxford, Parks Road, Oxford, UK}
\author{E. Viezzer}
\affiliation{Department of Atomic, Molecular and Nuclear Physics, University of Seville, 41012 Seville, Spain}
\affiliation{Max-Planck-Institut f\"{u}r Plasmaphysik, EURATOM Association, Boltzmannstr. 2, 85748 Garching, Germany}
\author{S. M. Vinko}
\affiliation{Clarendon Laboratory, Department of Physics, University of Oxford, Parks Road, Oxford, UK}

\date{\today}

\begin{abstract}
%\begin{linenumbers}

%\textit{Short summary}: Speeding up expensive simulations by up to 2 billion times by constructing accurate emulators, even when the datasets are limited.
%~
Computer simulations are invaluable tools for scientific discovery.
However, accurate simulations are often slow to execute, which limits their applicability to extensive parameter exploration, large-scale data analysis, and uncertainty quantification.
A promising route to accelerate simulations by building fast emulators with machine learning requires large training datasets, which can be prohibitively expensive to obtain with slow simulations.
Here we present a method based on neural architecture search to build accurate emulators even with a limited number of training data.
The method successfully accelerates simulations by up to 2 billion times in 10 scientific cases including astrophysics, climate science, biogeochemistry, high energy density physics, fusion energy, and seismology, using the same super-architecture, algorithm, and hyperparameters.
Our approach also inherently provides emulator uncertainty estimation, adding further confidence in their use.
We anticipate this work will accelerate research involving expensive simulations, allow more extensive parameters exploration, and enable new, previously unfeasible computational discovery.

%\end{linenumbers}
\end{abstract}

\maketitle

\section{Introduction}
Finding a general approach to speed up a large class of simulations would enable tasks that are otherwise prohibitively expensive and accelerate scientific research.
For example, fast and accurate simulations promise to speed up new materials and drug discovery~\cite{greeley2006computational-hts} by allowing rapid screening and ideas testing.
Accelerated simulations also open up novel possibilities for online diagnostics for cases like x-ray scattering in plasma physics experiments~\cite{lee-2009-thomson-scattering} and to monitor edge-localized modes in magnetic confinement fusion~\cite{galdon2018beam-elms}, enabling real-time prediction-based experimental control and optimization. However, for such applications to be successful the simulations need not only be fast but also accurate; achieving both to the level required for advanced applications remains an active objective of current research.

One popular approach to speeding up simulations is to train machine learning models to emulate slow simulations~\cite{peterson2017zonal,kwan2015cosmic-emu,brockherde2017bypassing,rupp2012fast-molecular-energy-krr} and use the emulators instead.
The main challenge in constructing emulators with machine learning models is in their need for large amounts of training data to achieve the required accuracy in replicating the outputs of the simulations. This training data could be prohibitively expensive to generate with slow simulations.

To construct high fidelity emulators with limited training data, the machine learning models need to have a good prior on the simulation models.
Most work to date in building emulators, using random forests~\cite{peterson2017zonal}, Gaussian Processes~\cite{kwan2015cosmic-emu}, or other machine learning models,~\cite{brockherde2017bypassing,rupp2012fast-molecular-energy-krr} do not fully capture the correlation among the output points, limiting their accuracy in emulating simulations with one, two, or three-dimensional output signals.
On the other hand, convolutional neural networks (CNN) have shown to have a good prior on natural signals~\cite{ulyanov2018deep-image-prior}, making them suitable for processing natural $n$-dimensional signals.
However, as the CNN priors inherently rely on their architectures~\cite{ulyanov2018deep-image-prior}, one has to find an architecture that gives the suitable prior for a given problem.
Manually searching for the right architecture can take a significant amount of time and domain-specific expertise, and often produces sub-optimal results.

%Most work to date in building emulators is based on using random forests~\cite{peterson2017zonal}, Gaussian Processes~\cite{kwan2015cosmic-emu}, or other machine learning models~\cite{brockherde2017bypassing,rupp2012fast-molecular-energy-krr}.
%However, the limitations of these models in learning complex relationships between the inputs and outputs of simulations restrict their prediction accuracy, making them useful only in some limited cases~\cite{kwan2015cosmic-emu,smith2017ani-anakinme}.

%An alternative solution is to employ deep neural networks to learn the complex input-output relations~\cite{smith2017ani-anakinme,anirudh2019cyclegan,montavon2012learning-molecular-energy-mlp}.
%However, significant effort is required to find the right deep neural network architecture for a given problem.
%Employing an architecture that is too complex or too simple for a problem makes it prone to overfitting or underfitting, both of which reduce the out-of-samples accuracy of the emulator.
%Manually searching for the right architecture for a given problem can take a significant amount of time and domain-specific expertise, but also often produces sub-optimal results.
%The problem of finding the right architecture is even more challenging with expensive simulations where only limited training data can be generated.

%\section{Deep Emulator Network SEarch}

Here we propose to address this problem by employing efficient neural architecture search~\cite{pham2018efficient-enas,cai2018proxylessnas} to simultaneously find the neural network architecture that is well-suited for a given case and train it.
With the efficient neural architecture search and a novel super-architecture presented in this work, the algorithm can find and train fast emulators for a wide range of applications while offering major improvements in terms of accuracy compared with other techniques, even when the training data is limited.
We call the presented method Deep Emulator Network SEarch (DENSE).

\begin{figure*}
    \centering
    \includegraphics[width=\linewidth]{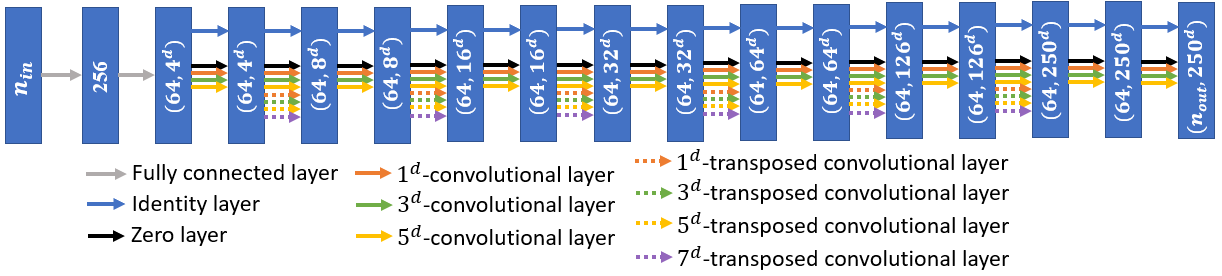}
    \caption{The super-architecture used in this paper where $d$ is the dimension of the output signal.
    The first numbers in the brackets indicate the number of channels and the last numbers indicate the signal size.
    For cases where the output signal are not $250^d$, then all the sizes in the intermediate nodes are scaled accordingly.
    Close arrows indicate the operations in the same group.
    The output of the identity layer and the selected of convolutional layer are added into the destination node.}
    \label{fig:super-architecture}
\end{figure*}

In DENSE, we start by defining the search space of neural network architectures in a form of super-architecture.
A super-architecture consists of multiple nodes where the first node represents the simulation inputs and the last node the predicted simulation outputs.
Each pair of nodes is connected by multiple groups of operations.
Each group consists of a set of operations, such as $1\times 1$ convolution, $3\times 3$ convolution, or similar.
Most of the operations, such as convolution, contain sets of trainable values that are commonly known as {\it weights}.
In one forward calculation of the neural network (i.e. predicting a set of outputs given some input), only one operation per group is chosen according to its assigned probability.
The probability of an operation being chosen is determined by a trainable value associated with the operation, which we call the {\it network variable}.

The super-architecture used in this work is shown in Figure \ref{fig:super-architecture}.
In every group in the super-architecture, there are convolutional layers with different kernel sizes and a zero layer that multiplies the input with zero.
The option of having a zero layer and multiple convolutional layers enables the algorithm to choose an appropriate architecture complexity for a given problem.
The super-architecture also contains skip connections~\cite{he2016deep-resnet} (i.e. identity layers) to make it easier to train.

Training the neural network involves two update steps.
In the first step, an operation for each group is chosen according to its probability, forming one possible architecture.
The weights, $\mathbf{w}$, of the selected operations are then updated to minimize the expected value of a defined loss function, $\mathcal{L}$, between the predicted simulation output and the actual simulation output,
\begin{equation}
    \label{eq:weights-update}
    \mathbf{w} \leftarrow \mathbf{w} - \alpha_1 \nabla_{\mathbf{w}}\mathbb{E}_{a\sim\mathcal{A}(\mathbf{b})}\left[\mathcal{L}(\mathbf{w} | \mathbf{X_t},\mathbf{y_t},a)\right],
\end{equation}
where $\alpha_1$ is the update step size, $\mathbf{X_t}$ and $\mathbf{y_t}$ are the input and output from the training dataset, $a$ is an architecture sampled from the super-architecture $\mathcal{A}(\mathbf{b})$ according to the network variables, $\mathbf{b}$.
The loss function in this paper is defined as the Huber loss function~\cite{huber1992robust-loss} to minimize the effect of outlier data and increase robustness.

The second update step involves evaluating the performance of various sampled architectures on the validation dataset, which is different from the training dataset employed in the first step.
The performance of an architecture can be evaluated based on the loss function, inference time, power consumption, or some other combination of relevant criteria.
The architectures are then ranked based on their performance and they are given rewards according to their rank.
The network variables, $\mathbf{b}$, are updated to increase the probability of the high-ranked architectures and decrease the probability of the low-ranked architectures.
Formally, the update can be written as~\cite{williams1992simple-REINFORCE},
\begin{equation}
    \label{eq:update-network-var}
    \mathbf{b} \leftarrow \mathbf{b} +
    \alpha_2 \mathbb{E}_{a\sim\mathcal{A}(\mathbf{b})}\left(\mathcal{R}_a\nabla_\mathbf{b}\log\left[\pi(a|\mathbf{b})\right]\right),
\end{equation}
where $\alpha_2$ is the update step size, $\mathcal{R}_a$ is the reward value given to the architecture $a$ based on its rank, and the function $\pi(a|\mathbf{b})$ is the likelihood of the architecture $a$ being chosen given the network variables $\mathbf{b}$.

In this case, we ranked the architectures based on the Huber loss~\cite{huber1992robust-loss} on the validation dataset and gave the rewards to follow the zero-mean ranking function in CMA-ES~\cite{hansen2016-cmaes}.
The use of a zero-mean ranking function reduces the update variance and makes the update step scale-invariant, increasing the robustness of the algorithm.

\begin{figure*}
    \centering
    \includegraphics[width=\linewidth]{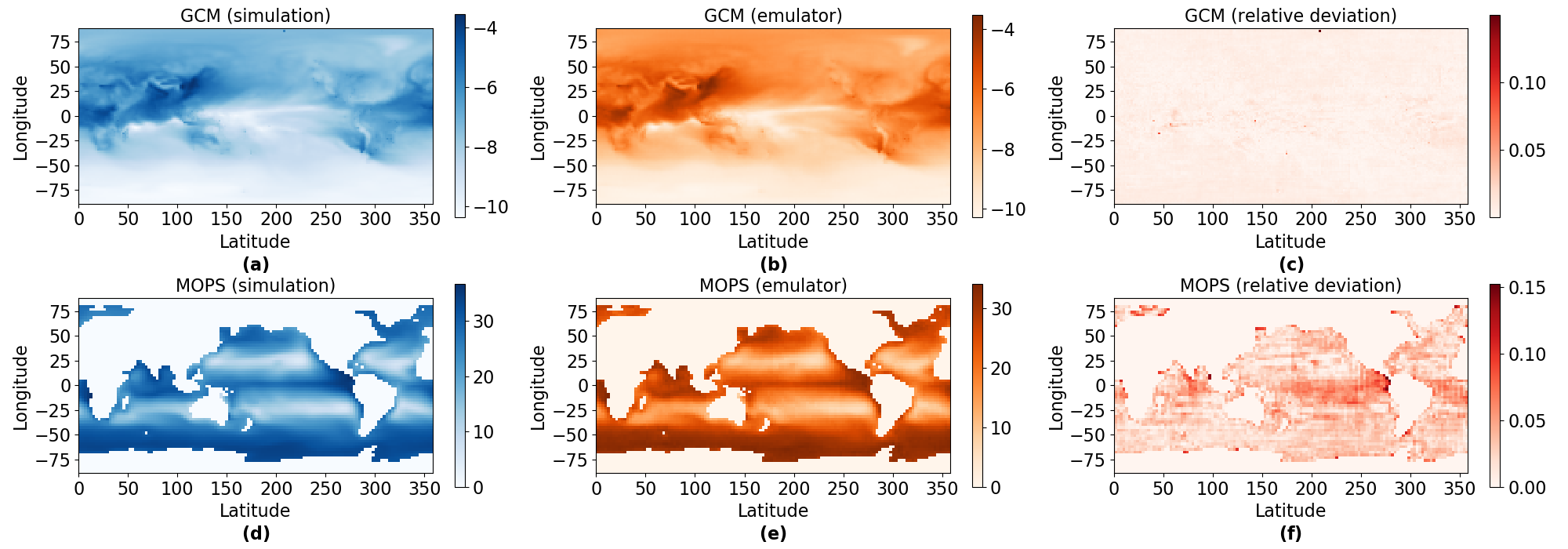}
    \caption{Original simulations (a,d) compared with outputs from their DENSE emulators (b,e) on a representative example from the test dataset. \revision{The relative differences are shown in (c,f)}. Outputs from the remaining 8 cases are given in the methods section in Figure \ref{fig:emulator-results-all-test-cases}.
    }
    \label{fig:emulator-results}
\end{figure*}

\begin{figure*}
    \centering
    \includegraphics[width=\linewidth]{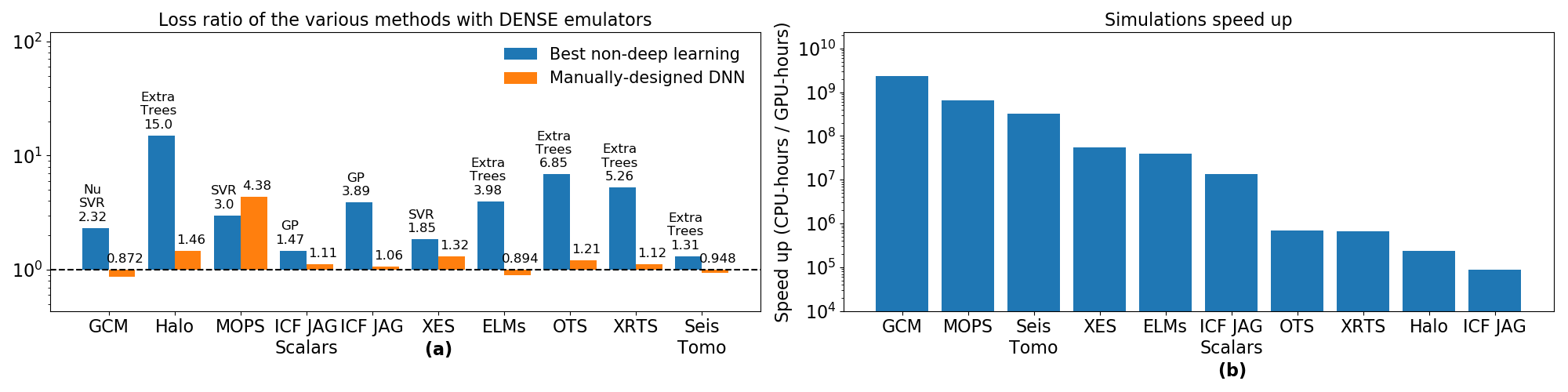}
    \caption{
    (a) The ratio between the loss function obtained by DENSE emulators and the best loss function found by non-deep learning methods and a manually-designed deep neural network.
    (b) The achieved speed up of the emulators using a GPU compared with the original simulations.}
    \label{fig:accuracy-speedup-results}
\end{figure*}

\begin{figure*}
    \centering
    \includegraphics[width=\linewidth]{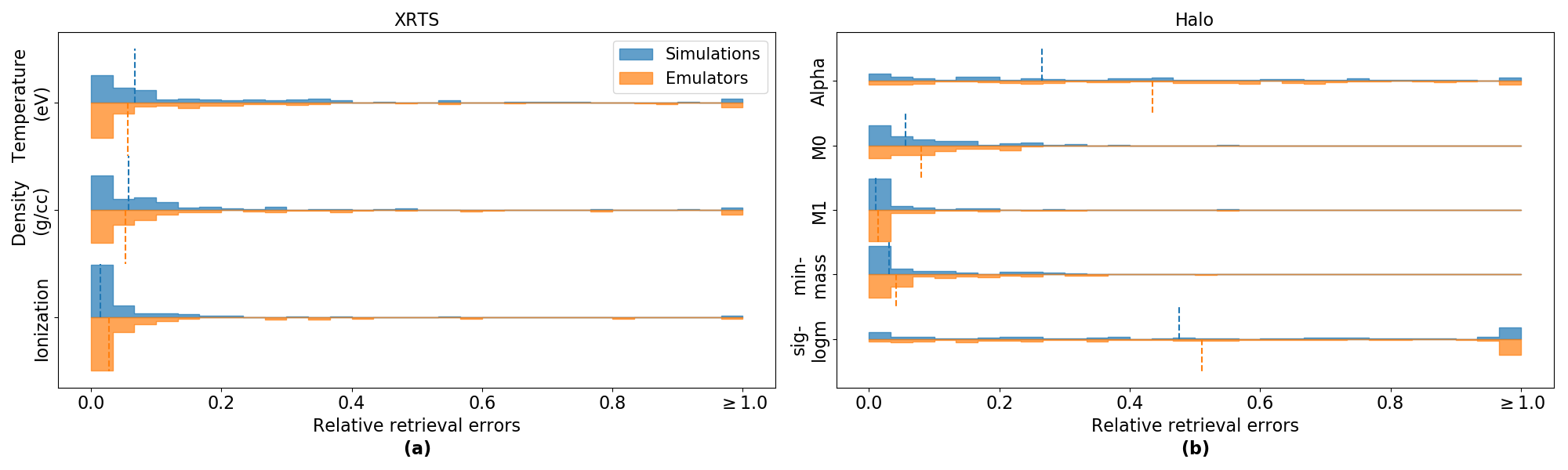}
    \caption{
    Histograms of the relative retrieval errors in solving the inverse problem with noisy data for the XRTS (a) and Halo (b) test cases. Retrieval conducted using the full simulations is shown in blue, and the retrieval using DENSE emulators in orange. Median values of the distributions are shown with the dashed lines.}
    \label{fig:inv-opt-results}
\end{figure*}
\begin{figure*}
    \centering
    \includegraphics[width=\linewidth]{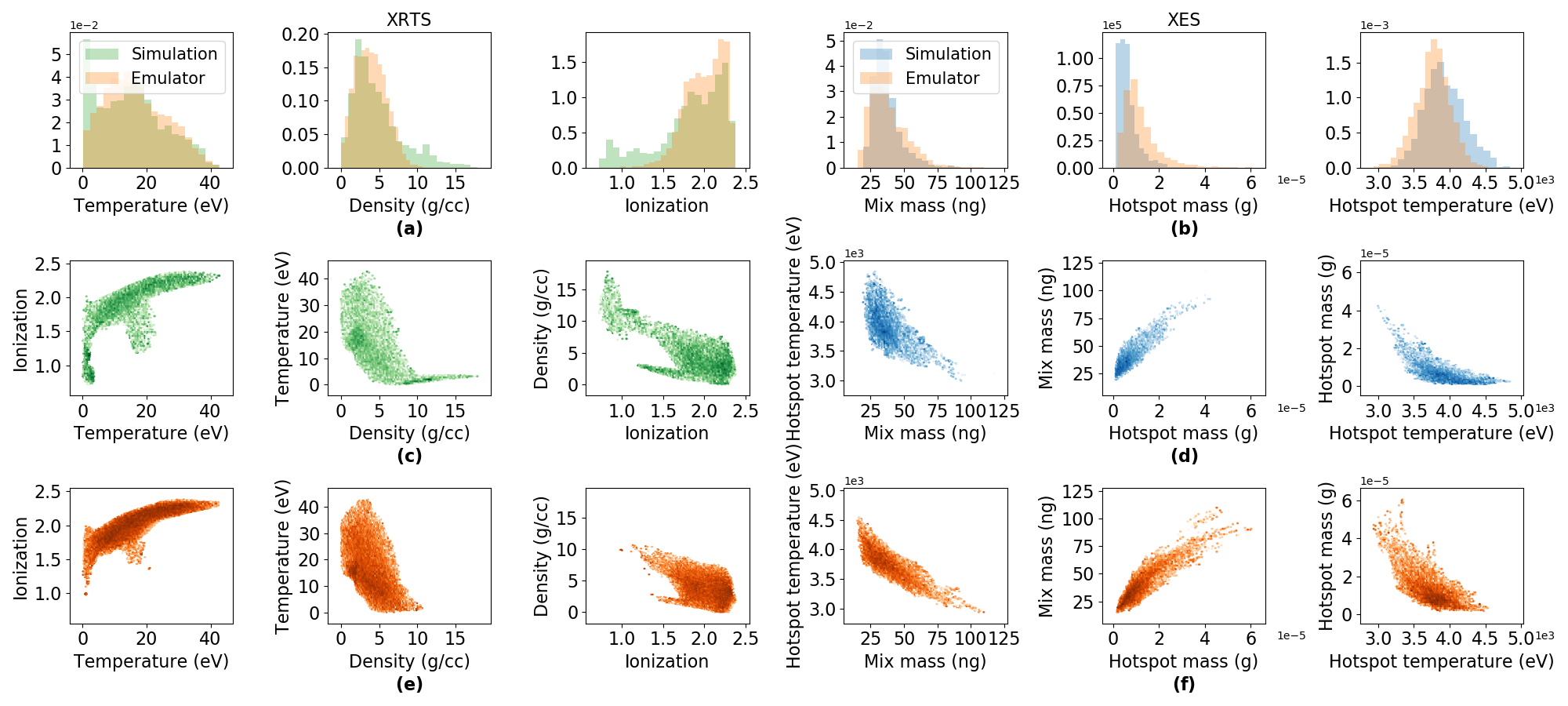}
    \caption{
    Comparison of results of Bayesian posterior sampling when applied using simulations and emulators in solving inverse problems. Histograms for the posterior distributions are shown for three retrieved parameters in the XES (a) and XRTS (b) cases. The parameter posterior distribution scatter plots are shown in panels (c,d) for the simulations, and (e,f) for the DENSE emulators.}
    \label{fig:inv-mcmc-results}
\end{figure*}

\begin{figure*}
    \centering
    \includegraphics[width=\linewidth]{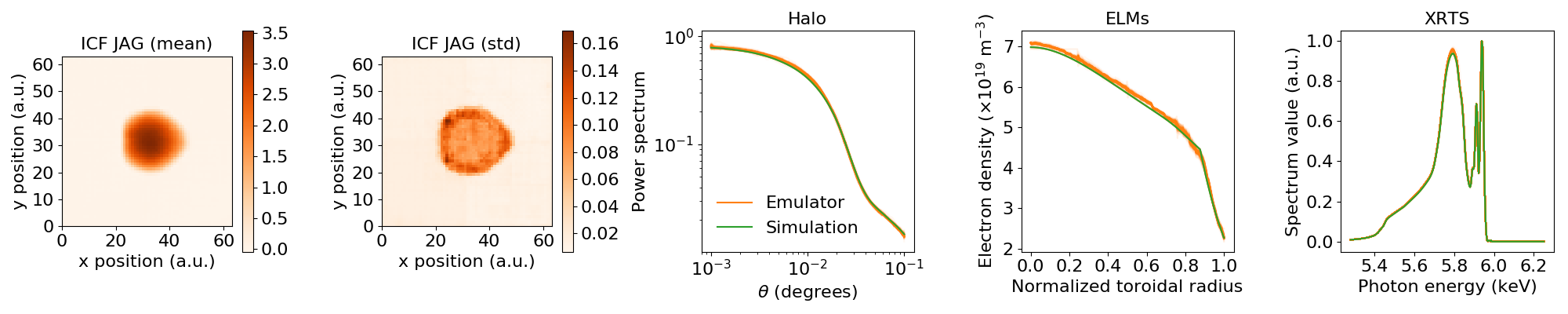}
    \caption{
    Emulator prediction uncertainties: comparison of DENSE emulator predictions with outputs from simulations for 2D (ICF JAG) and 1D models (Halo, ELMs, and XRTS).
    \revision{The orange lines in Halo, ELMs, and XRTS resemble the distribution of the emulators outputs while the green lines are the output from the simulations.}}
    \label{fig:uncertainty-results}
\end{figure*}

\section{Results}

The combined update steps from equations (\ref{eq:weights-update}) and (\ref{eq:update-network-var}), and the use of a ranking function in assigning rewards, make DENSE a robust algorithm to simultaneously learn the weights and find the right architecture for a given problem.
To illustrate this, we apply the method to ten distinct scientific simulation cases:
inelastic x-ray Thomson scattering (XRTS) in high-energy-density physics~\cite{gregori2003theoretical-thomson-scattering, lee-2009-thomson-scattering},
optical Thomson scattering (OTS) in laboratory astrophysics~\cite{tzeferacos2018laboratory-ots},
tokamak edge-localised modes diagnostics (ELMs) in fusion energy science~\cite{galdon2018beam-elms},
x-ray emission spectroscopy (XES) in plasmas~\cite{regan2013hot-xes, ciricosta2017simultaneous-xes},
galaxy halo occupation distribution modelling (Halo) in astrophysics~\cite{hatfield2016galaxy-halo},
seismic tomography of the Shatsky Rise oceanic plateau (SeisTomo)~\cite{korenaga2012seismic-mcmc},
global aerosol-climate modelling using a general circulation model (GCM) in climate science~\cite{tegen2019global-climate-model-aerosol},
oceanic pelagic stoichiometry modelling (MOPS) in biogeochemistry~\cite{khatiwala2007computational},
and
neutron imaging (ICF JAG) and scalar measurements (ICF JAG Scalars) in inertial confinement fusion experiments~\cite{anirudh2019cyclegan}.

The tested simulations have ranging numbers of scalar input parameters from 3 to 14, and span outputs from 0D (scalars) to multiple \revision{2}D signals (images).
%The output of the simulations are between 4 and 12 two-dimensional signals (images) for ICF JAG and GCM, 15 scalar values for ICF JAG Scalars, and between 1 and 10 one-dimensional signals (functions) for the rest.
Datasets for simulations that run in less than 1 CPU-hour were generated by running them 14,000 times with random sets of inputs.
For more expensive simulations, the number of generated dataset is limited by the time and resources available (see table \ref{tab:testcase-table} for more detailed information).
Each dataset is divided into three parts: 50\% is used as the training dataset, 21\% for validation, and 29\% as the test dataset.
%Every test case has a dataset which consist of 39 to 14,000 data points where half of it is used as the training dataset, 21\% as the validation dataset, and 29\% as the test dataset.
The test dataset was used only to present the results in this paper, never to build the models.
%\revision{The output predictions of the test dataset was done using the architecture with the highest likelihood.}
The hyperparameters were obtained by optimizing the result for OTS with CMA-ES~\cite{hansen2016-cmaes,loshchilov2016-cmaes-hyperparameters-tuning}, then used for other cases without further tuning.

\subsection{Emulator results}

The example outputs of the trained emulators with DENSE are shown in Figure \ref{fig:emulator-results}.
%\revision{The images were generated using the architecture with the highest likelihood.}
We see that the output of the emulators generally matches closely the output of the actual simulations, even in MOPS and GCM where only 410 and 39 data points are available.
When only a limited number of data is available, the choice of model architectures that give the right priors become important.
Complex model architectures with bad priors could still fit the sampled training data, but are likely to overfit, giving bad accuracy on the out-of-samples data.

With DENSE, the model architecture that gives a good prior on the problem is automatically searched for by preferring models that can fit well the out-of-samples data (i.e. the validation dataset).
Moreover, randomly choosing an operation in every layer acts as a regularizer in updating the weights during the training to minimize overfitting.
These two advantages make DENSE suitable for learning to emulate a wide range of simulations including expensive ones where only a limited number of datasets can be generated.

While the simulations presented typically run in minutes to days, the DENSE emulators can process multiple sets of input parameters in milliseconds to a few seconds with one CPU core, or even faster when using a GPU card.
For the GCM simulation which takes about 1150 CPU-hours to run, the emulator speedup is a factor of 110 million on a like-for-like basis, and over 2 billion with a GPU card.
The speed up achieved by DENSE emulators for each test case is shown in Figure \ref{fig:accuracy-speedup-results}(b).

Compared with other non-deep learning techniques usually employed in building emulators~\cite{pedregosa2011scikit}, the models found and trained by DENSE achieved the best results in all tested cases, and in most cases by a significant margin.
The DENSE model also performs better than an emulator designed specifically for ICF JAG simulation~\cite{anirudh2020improved} (i.e. CycleGAN in Figure 1 of the supplementary material).
As seen in Figure~\ref{fig:accuracy-speedup-results}(a), the emulators built by DENSE achieved a loss function up to 14 times lower than the best performing non-deep learning model.

We also compared DENSE with a manually-designed deep neural network model by an architecture from the super-architecture in Figure \ref{fig:super-architecture}, where all the convolutional layers have size 3.
The use of kernel size 3 and skip connections follows the idea of ResNet~\cite{he2016deep-resnet}.
As with DENSE, the hyperparameters for manually-designed deep neural network were tuned to optimize its result for OTS.

Shown in Figure \ref{fig:accuracy-speedup-results}(a) is the comparison between DENSE emulators and manually-designed deep neural network.
Although in most cases their performances are similar, in some cases DENSE emulators can give a considerable improvement in terms of loss function, as can be seen in Halo and MOPS.
This illustrates the robustness of DENSE emulators in a wide range of cases.

\subsection{\label{subsec:inverse-problem}Emulators for inverse problems}

The high fidelity emulators built by DENSE are sufficiently accurate to allow us to substitute simulations even for more advanced tasks such as solving the inverse problem~\cite{kasim-18-ipi}.
To illustrate this, we took a simulated output signal randomly from the test dataset where the actual parameters are known.
A small noise ($\sim 1\%$) was added to the chosen signal to closely mimic a real observed signal from an experiment.
Using this signal, we use an optimization algorithm~\cite{wierstra2008-natural-nes} to retrieve the input parameters by minimizing the error between the sample signal and the output of the emulators, which we call it the \textit{retrieval error}.

The results of the parameter retrieval using the emulators are compared with the retrieval using the simulations in Figures~\ref{fig:inv-opt-results}, where we plot the relative retrieval error histograms for two cases.
We observe that the relative retrieval errors from the emulators are very similar to those from the simulations.
%For some parameters, we see relatively flat and large relative error histograms such as~\textit{sig-logm} in the Halo test case, even when retrieved using the actual simulation.
%This is due to the ill-posedness of the problem where small noise in the observable maps onto a large deviation in the space of parameters~\cite{kasim-18-ipi}.

Without much loss in accuracy, the parameter retrieval with the emulators only takes about 800 ms with a single GPU card.
This is to be compared with using the actual simulations which could take up to 2 days (XES) even when using 32 CPU cores.
As the parameters can be retrieved in less than one second rather than in hours or days, one can envisage employing this technique for online diagnostics, real-time data interpretation, or even on-the-fly intelligent automation with an accuracy comparable to high-fidelity simulations that are by far too computationally expensive to be used directly.
The use of DENSE emulators also enables parameter retrieval with resource-intensive simulations, such as MOPS and GCM, that were too expensive before.

In addition to interpreting signals and parameter retrieval, the emulators can also be used to significantly speed up Bayesian uncertainty quantification~\cite{kasim-18-ipi}.
Bayesian uncertainty quantification is usually done by constructing Bayesian posterior distribution using Markov Chain Monte Carlo (MCMC) algorithms.
However, the cost of running MCMC to collect sufficient samples from the Bayesian posterior distribution is typically much larger than the cost for parameter retrieval, and is often intractable in practice.
Here we perform the Bayesian posterior sampling using an ensemble MCMC algorithm~\cite{goodman2010-ensemble-mcmc} with the same conditions as in ref~\cite{kasim-18-ipi}.
In short, we collect all parameters sets that produce spectra that lie in a certain band around a central spectrum.

Figure~\ref{fig:inv-mcmc-results} compares the results of sampling the posterior distribution using simulations and emulators in two cases to interpret scattering and spectroscopy data.
The posterior distribution sampled by the emulators are very similar to those by actual simulations, and we see that the emulators are well-suited to capture the correlations between parameters.
However, note that while collecting 200,000 XES samples via simulations takes over 22 days, the sampling process with the emulators was completed in just a few seconds.
Interestingly, building the emulator for XES from scratch only needs some 14,000 samples plus 8 hours for training, so the whole pipeline to build the emulator and use it for MCMC is still considerably faster than directly collecting 200,000 samples using the original simulation.

%From Fig.~\ref{fig:inv-results} we note some slight deviations in the distributions between the samples obtained using the simulations and the emulators.
%This deviation arises from the edge cases where the spectra lie at the edges of the acceptance band.
%Nevertheless, the sampling process using the emulators can still recover the main shape of the distribution and the statistics of the uncertainty.

\subsection{\label{subsec:prediction-uncertainty}Prediction uncertainty}

A final important advantage of building emulators with DENSE is the availability of an intrinsic estimator of the emulator's prediction uncertainty.
The randomization of network architectures from the super-architecture can be seen as a special case of dropout \cite{srivastava2014dropout}.
Thus, by adapting the theory of prediction uncertainty with Monte Carlo (MC) dropout by ref. \cite{gal2016dropout}, we can show that DENSE emulators can produce the uncertainty of their outputs.
The expected value and variance of  a DENSE emulator prediction can be obtained by

\begin{align}
\begin{aligned}
    \mathbb{E}(\mathbf{y}|\mathbf{x}) &= \mathbb{E}_{a\sim\mathcal{A}(\mathbf{b})}\left(\mathbf{y} | \mathbf{x}, a\right)\\
    \mathrm{Var}(\mathbf{y}|\mathbf{x}) &= \mathrm{Var}_{a\sim\mathcal{A}(\mathbf{b})}\left(\mathbf{y} | \mathbf{x}, a\right),
\end{aligned}
\end{align}
where $a$ is the architecture sampled from the super-architecture $\mathcal{A}$ based on the final values of the network parameters, $\mathbf{b}$.
Figure~\ref{fig:uncertainty-results} shows the prediction uncertainty of the DENSE emulators, illustrating regions where they are either uncertain or confident in their predictions.

\revision{As also observed in MC dropout,} \cite{gal2017concrete} \revision{the prediction uncertainty in this case can be smaller than the difference between the predicted and simulated output, indicating an overconfident prediction.
This problem of overconfidence can be resolved by stopping the training of network variables early.
The investigation of prediction uncertainty tuning will be the subject of future work.}

\section{Discussions}

\subsection{\revision{Limitations}}

Although DENSE has the capability of emulating a wide range of simulations, it is still limited to simulations with a few scalar inputs.
The DENSE algorithm has not been tested on building emulators with one, two, or three-dimensional direct inputs.
One way to fit a simulation with multi-dimensional inputs to DENSE is by parameterizing the inputs using several scalar parameters (as done in ELMs) or employing a dimensionality reduction techniques~\cite{mcinnes2018umap}.

Another limitation observed in DENSE is that it does not learn very well in regions where output variability is high, i.e. where changing the input parameters slightly changes the outputs significantly.
This limitation is also observed in other cases of deep learning~\cite{ronen2019convergence}.
Due to the difficulty in learning, regions with high variability tend to require more samples than regions with low variability.
This problem can thus be overcome by sampling the parameter space intelligently~\cite{chowdhury2019efficient}.

\subsection{\revision{Applications}}

Our DENSE approach opens up numerous applications that require fast calculations.
One of the main applications of DENSE emulators is real-time diagnostics of complex systems.
For research in some fields, such as in plasma physics, diagnostics are usually done by solving an inverse problem using simulations, which involves running the simulations hundreds of times or more.
By using the fast emulators instead of simulations, solving the inverse problem could be significantly accelerated without sacrificing the quality of the solutions, as shown in section \ref{subsec:inverse-problem}.
Real-time diagnostics are the key in automating operations of some machines with complex systems, such as tokamaks \cite{kates2019predicting-tokamak} and particle accelerators \cite{emma2018machine-learning-particle-accelerator}.

Another application is the optimization of a very expensive simulations where the simulations can only be executed a few times.
The emulators can be used to make high-quality guesses about the optimum parameters which can then be tested using the expensive simulations.
This idea of utilizing a cheap model for expensive simulation optimization has been used in surrogate-model optimization \cite{liu2013gp-surrogate-optimization} and Bayesian optimization \cite{shahriari2015bayesopt}.
By using a more accurate cheap model, such as DENSE emulators, the number of simulations to be executed can be much lower than in previous approaches.
This is a potential avenue for future works.

\section{Conclusions}
We have shown that Deep Emulator Network SEarch (DENSE), a method based on neural architecture search, can be used to robustly build fast and accurate emulators for various types of scientific simulations even with limited training data.
The capability of DENSE to accurately emulate simulations with limited data makes the acceleration of very expensive simulations possible.
With the achieved acceleration of up to 2 billion times, DENSE emulators enable tasks that were impossible before, such as real-time simulation-based diagnostics, uncertainty quantification, and extensive parameters exploration.
This large acceleration in solving inverse problems removes the barriers of using high fidelity simulations in real-time measurements, opening up new types of online diagnostics in the future.
The wide range of successful test cases presented here shows the generality of the method in speeding up simulations, enabling rapid idea testing and accelerating new discovery across the sciences and engineering.

%The ability to build accurate and fast emulators for a large class of simulations as presented in this paper enables rapid ideas testing on large search spaces, accelerating new discovery across the sciences and engineering.

\section{\label{sec:method} Methods}

\subsection{\label{subsec:test-cases} Test cases}

\begin{table*}[t!]
\begin{center}
\begin{tabular}{ |c|c|c|c|c|c|c| } 
 \hline
 \textbf{No.} &
 \textbf{Test case} & \textbf{\# Inputs} & \textbf{\# Outputs} & \textbf{Output type} & \textbf{\# Dataset} & \textbf{Avg. simulation running time}\\
 \hline
 1 & XRTS & 3 & 1 & 1D (250 points) & 14,000 & 15 seconds \\
 2 & OTS & 5 & 1 & 1D (250 points) & 14,000 & 15 seconds \\
 3 & XES & 10 & 1 & 1D (250 points) & 14,000 & 20 minutes \\
 4 & ELMs & 14 & 10 & 1D (250 points) & 14,000 & 15 minutes \\
 5 & Halo & 5 & 1 & 1D (250 points) & 14,000 & 5 seconds\\
 6 & ICF JAG & 5 & 4 & 2D ($64\times64$) & 10,000 & 30 seconds \\
 7 & ICF JAG Scalars & 5 & 15 & 0D (scalar) & 10,000 & 30 seconds \\
 8 & SeisTomo & 13 & 1 & 1D (250 points) & 6,100 & 2 hours \\
 9 & MOPS & 6 & 45 & 2D ($128\times 64$) & 410 & 144 CPU-hours \\
 10 & GCM & 3 & 12 & 2D ($192\times96$) & 39 & 1150 CPU-hours \\
 \hline
\end{tabular}
\end{center}
\caption{Summary of test cases considered in this paper. The inputs to all simulations are all 0D (scalars). The number of datasets for relatively fast simulations is capped to 14,000 for convenience. For slower simulations the size of the dataset is limited by time and resource constraints.}
\label{tab:testcase-table}
\end{table*}

\begin{figure*}
    \centering
    \includegraphics[width=\linewidth]{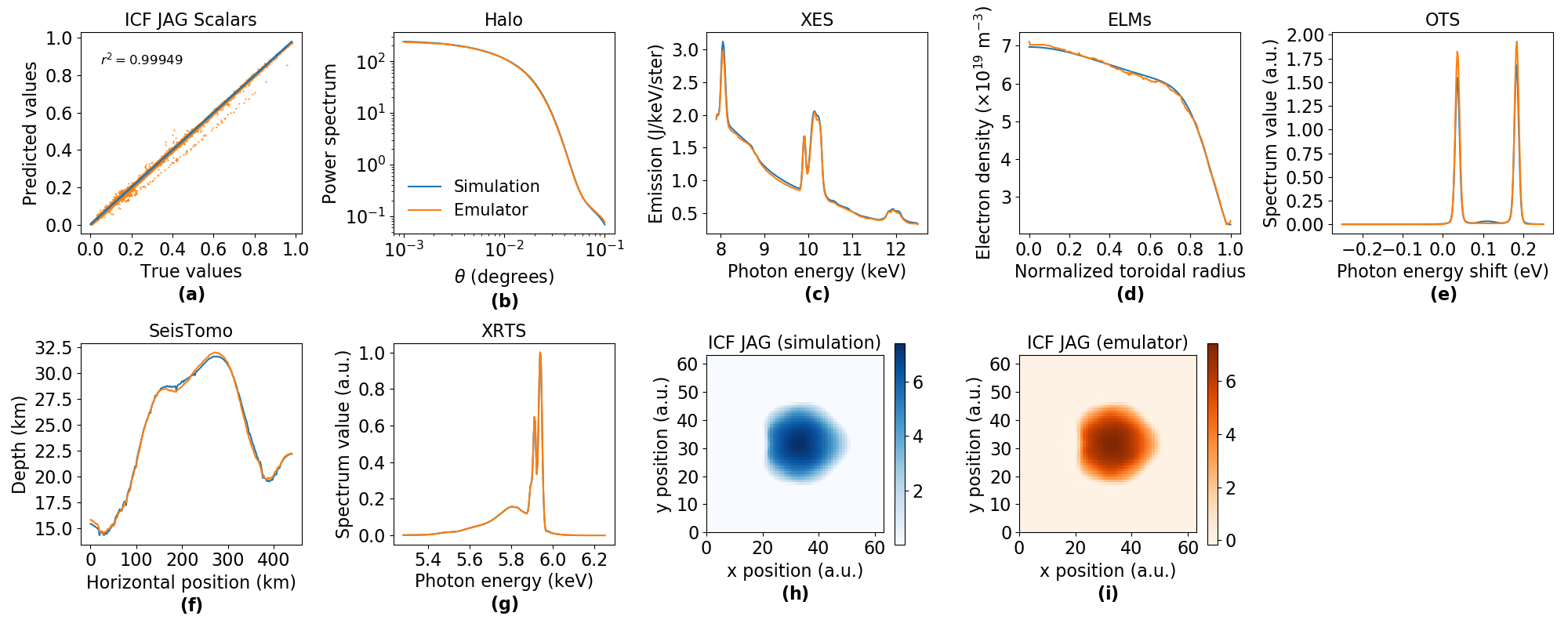}
    \caption{(a-i) Examples of emulator outputs for the remaining test cases not shown on Figure \ref{fig:emulator-results}.}
    \label{fig:emulator-results-all-test-cases}
\end{figure*}

\begin{figure*}
    \centering
    \includegraphics[width=\linewidth]{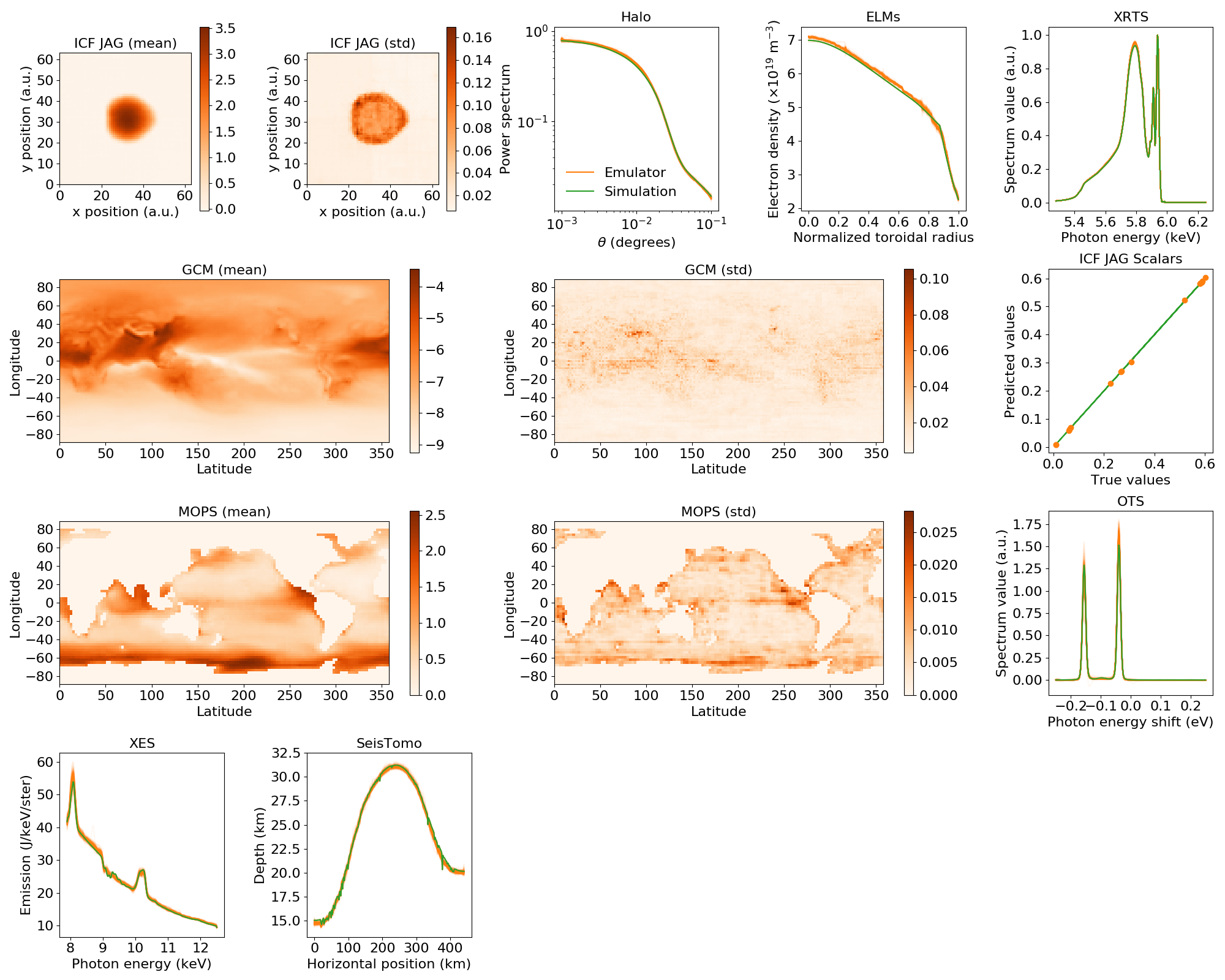}
    \caption{Examples of emulator uncertainties for test cases not shown in Figure \ref{fig:uncertainty-results}.
    The orange lines in 1D simulations resemble the distribution of the emulators outputs while the green lines are the outputs from the simulations.}
    \label{fig:emulator-uncertainty-all-test-cases}
\end{figure*}

Here we provide a description of the test cases employed in the paper.
A summary of the test case parameters is given in Table \ref{tab:testcase-table}.
All 1D simulation outputs are sampled to 250 points for convenience.

\textbf{X-ray Thomson scattering (XRTS)}:
XRTS is a technique widely used in high-energy-density physics to extract plasma temperatures and densities by measuring the spectrum of an inelastically scattered x-ray pulse~\cite{lee-2009-thomson-scattering, kritcher-2008-thomson-scattering}.
The spectrum of the scattered light can be calculated from a set of plasma conditions and the scattering geometry~\cite{gregori2003theoretical-thomson-scattering}; this forms the simulation on which our emulator is based.

In this paper we consider the specific experimental case presented in ref.~\cite{lee-2009-thomson-scattering} where three parameters (temperature, ionization, and density) are to be retrieved from a spectrum of x-rays scattered at a 90-degree angle from a shock-compressed Beryllium plasma. The high-speed emulator for XRTS enables fast solutions to the inverse problem and access to statistical information on the intrinsic uncertainty of the experiment, allowing better control of the experimental optimization and information extraction.

\textbf{Optical Thomson scattering (OTS)}:
OTS is conceptually similar to XRTS except that it uses optical light instead of x-rays.
Optical Thomson scattering is used in measuring electrons and ions temperatures and densities, as well as the flow speed of the plasma using the Doppler shift \cite{tzeferacos2018laboratory-ots}.

Here we considered retrieving five physical parameters (electron and ion temperatures, electron density, ionization, and flow speed) from a normalized scattered spectrum.
The impact of building an emulator for OTS is similar to XRTS as it enables access to real-time data interpretation and to uncertainty quantification.

\textbf{X-ray emission spectroscopy (XES)}:
X-ray emission spectroscopy is a general technique to probe a system by measuring the emitted spectrum and matching it with simulations or theoretical models. In this paper, we consider the diagnostic case of a laser-driven implosion experiment at the National Ignition Facility~\cite{regan2013hot-xes}, using the spectroscopic model based on the CRETIN atomic kinetics code described in detail in ref~\cite{ciricosta2017simultaneous-xes}.

\textbf{Edge-Localized Modes (ELMs) diagnostics}:
Edge-localized modes are magnetohydrodynamic instabilities that occur in magnetically confined fusion plasmas with high confinement \cite{zohm1996edge-elms}.
ELMs are explosive events and cause detrimental heat and particle loads on the plasma facing components of a tokamak.
Various diagnostics are implemented to track ELMs \cite{cavedon2017pedestal-elms}.
Here, we compare the emulator to the predictive model \cite{viezzer2018ion-elms-astra} for the temporal evolution of the electron density profile using the transport code ASTRA \cite{fable2013novel-elms-astra}.

The 14 input parameters in this case describe the diffusion, convective velocity, and particle source profiles \cite{willensdorfer2013particle-elms} as a function of toroidal radial position and time.
The output observable is the time-dependent electron density as a function of toroidal radial position.

\textbf{Galaxy halo occupation distribution modelling (Halo)}:
Here we considered simulations of the angular-scale correlation of a galaxy population.
The simulation software Halomod~\cite{halomodsoftware} was used to calculate the correlation function, angular scales, redshifts and the cosmological model as described in ref~\cite{hatfield2016galaxy-halo}.
The input parameters used in this simulation follows the parameters described in ref\cite{wake2011galaxy}.
Fast parameter retrieval is of particular interest here as often researchers are interested in extracting parameters for multiple different galaxy populations.

\textbf{JAG model for Inertial Confinement Fusion (ICF JAG)}:
JAG simulates the observables from an inertial confinement fusion experiment~\cite{anirudh2019cyclegan}.
There are 5 input parameters in the case considered here.
One simulation with 5 input parameters produces four two-dimensional images and 15 scalar values.
Constructing fast and accurate emulators of the model allows for a more efficient exploration of the parameters space, and to obtain optimum sets of parameters more efficiently.

\textbf{Shatsky Rise seismic tomography (SeisTomo)}:
The case considered here is the seismic tomographic inversion problem of the Shatsky Rise oceanic plateau~\cite{korenaga2012seismic-mcmc}.
Given the input parameters that describe the initial velocity profile and regularization in the optimization, the software solves for the velocity structure and the crustal thicknesses as a function of position in the Shatsky Rise that matches the seismic reflection data.
Performing uncertainty quantification of the tomographic inversion would require the execution of the software hundreds of thousand times which is very expensive without a fast emulator model.

\textbf{Global aerosol-climate modelling (GCM)}:
The model considered here is ECHAM-HAM~\cite{tegen2019global-climate-model-aerosol} which calculates the distribution and evolution of both internally and externally mixed aerosol species in the atmosphere and their affect on both radiation and cloud processes.
The model simulates the aerosol absorption optical depth as the observable for every month in a year.
The model receives three input parameters which are (1) a scaling of the emissions flux of Black Carbon (BC; the main absorbing aerosol species), (2) a scaling on the removal rate of BC through wet deposition, and (3) a scaling of the imaginary refractive index of BC (which determines it’s absorptivity) between 0.2 and 0.8. All of these factors contribute to the different absorption aerosol optical depths we emulate.

The cost of running the model for one year (including three months of spin-up) is about 1150 CPU-hours which is prohibitively expensive when generating thousands of training data points.
However, we have shown that an accurate emulator over three parameters can be built with as few as 39 data points.

\textbf{The Model of Oceanic Pelagic Stoichiometry (MOPS)}:
The Model of Oceanic Pelagic Stoichiometry (MOPS) is a global ocean biogeochemical model~\cite{kriest2015mops} that simulates the cycling of nutrients (i.e., phosphorus, nitrogen), phytoplankton, zooplankton, dissolved oxygen and dissolved inorganic carbon.
MOPS is coupled to the Transport Matrix Method (TMM), a computational framework for efficient advective-diffusive transport of ocean biogeochemical tracers~\cite{khatiwala2007computational,samarkhatiwala_2018_computational}.
In this study we use monthly mean transport matrices derived from a configuration of MITgcm~\cite{marshall1997finite-mitgcm} with a horizontal resolution of 2.8$^\circ$ and 15 vertical levels.
There are 6 MOPS input parameters considered in this case, whose definitions and ranges are described in an optimization study by ref~\cite{kriest2017calibrating-mops}.
Each simulation involves integrating the model for 3000 years to a steady state starting from a uniform spatial distribution of tracers.
Annual mean 3D fields of oxygen, phosphorus, and nitrate at the end of the simulation are used for training.
All code and relevant data used for the simulations are freely available~\cite{samarkhatiwala_2018_computational}.

\subsection{\label{subsec:super-architecture} Super-architecture}

The super-architecture employed for most cases is shown in Figure \ref{fig:super-architecture}.
It consists of two fully connected layers at the beginning, followed by combinations of different types of convolutional layers.
Rectified Linear Units are used as the nonlinearity.
Most of the nodes contain 64 channels with a growing size of the signal from 4 to 250 at the end.
For ICF JAG that has output signal of size $64\times 64$, the size written in Figure \ref{fig:super-architecture} is capped at 64.

After the first two fully connected layers, each pair of adjacent nodes are connected by an identity layer and a selection of multiple convolutional layers.
The identity layers serve as the skip connection for each layer which is always present in every sampled architecture.
The member $j$ in group $i$ of layers is assigned a network parameters, $b_{ij}$, and the probability of member $j$ being selected among the group is determined by the softmax function,
\begin{equation}
\label{eq:prob-equation}
    p_{ij} = \frac{\exp{b_{ij}}}{\sum_k \exp{b_{ik}}}.
\end{equation}

Each group of layers consist of one zero layer which multiplies all the inputs to zero.
This provides an option for DENSE to remove the layer and make the neural network shallower.

For two nodes with different sizes, we added modified transposed convolutional layers in the group.
In the modified transposed convolutional layers, the signal is expanded just as in the normal transposed convolutional layer, but the ``holes'' are filled with a trainable constant instead of zeros.
The convolutional layers and identity layers between two nodes of different sizes operate by upsampling the previous node to match the size of the target node using the nearest neighbor.

\subsection{\label{subsec:training-pseudocode} Hyperparameters}
%\subsection{\label{subsec:training-pseudocode} Algorithm pseudo-code and hyperparameters}

%The pseudo-code of the DENSE training algorithm is shown in Algorithm \ref{alg:pseudocode} with Huber loss function is defined as
%\begin{equation}
%    \mathrm{Huber}(x) = \begin{cases}
%    \frac{1}{2} x^2; & |x| \leq 1 \\
%    |x| - \frac{1}{2}; & |x| > 1.
%    \end{cases}
%\end{equation}
%The use of Huber loss function increase the robustness of the training in the presence of outlier data without having to remove them.
The list of hyperparameters used in the algorithm are given in Table \ref{tab:hparams-table}.
The hyperparameters were chosen to minimize the validation loss function for OTS using CMA-ES~\cite{loshchilov2016-cmaes-hyperparameters-tuning, hansen2016-cmaes}.
The same hyperparameters were used for all cases.

%\begin{algorithm}[t!]
%\caption{DENSE training algorithm}
%\label{alg:pseudocode}
%\begin{algorithmic}[1]
%\State Initialize $\mathbf{w}$ and $\mathbf{b}$
%\For{$e = 1$ to $n_{epochs}$}
%    \State \textit{\% The first update step}
%    \For{$g = 1$ to $\lceil n_{train} / m_1\rceil$}
%        \State $\mathbf{X_t}, \mathbf{y_t} \leftarrow$ $m_1$ rows from the training dataset
%        \State $a \leftarrow$ a randomly chosen architecture
%        \State $\mathcal{L}_t = \mathrm{Huber}(a(\mathbf{X_t}, \mathbf{w}) - \mathbf{y_t})$
%        \State $\mathbf{w} \leftarrow \mathbf{w} - \alpha_1 \nabla_{\mathbf{w}}\mathcal{L}_t$
%    \EndFor
%    \State ~
%    \State \textit{\% The second update step}
%    \State \textit{\% The validation dataset iterated through $p_{val}$ times}
%    \State $\mu = p_{val} \lceil n_{val} / m_2\rceil$
%    \For{$k = 1$ to $\mu$}
%        \State $\mathbf{X_v}, \mathbf{y_v} \leftarrow$ $m_2$ rows from the validation dataset
%        \State $a_k \leftarrow$ a randomly chosen architecture
%        \State $\pi_k = \prod_{\mathrm{selected}(i,j)} p_{ij}(\mathbf{b})$ from Eq. (\ref{eq:prob-equation})
%        \State $\mathcal{L}_k = \mathrm{Huber}(a_k(\mathbf{X_v}, \mathbf{w}) - \mathbf{y_v})$
%    \EndFor
%    \State Sort $a_k$ with increasing $\mathcal{L}_k$ to get the ranks, $r_k$.
%    \State Give reward $\mathcal{R}_k = \frac{\log(\mu+1/2) - \log(r_k)}{\mu \log(\mu+1/2) - \sum_q \log(r_q)} - 1/\mu$
%    \State $\mathbf{b} \leftarrow \mathbf{b} + \alpha_2 \sum_k \mathcal{R}_k\nabla_\mathbf{b}\log(\pi_k)$
%\EndFor
%\end{algorithmic}
%\end{algorithm}

\begin{table*}[t!]
\begin{center}
\begin{tabular}{ |c|c|c|m{6.0cm}| } 
 \hline
 \textbf{Hyperparameters} & \textbf{For DENSE} & \textbf{For manual design DNN} & \textbf{Notes} \\ 
 \hline
 $n_{epochs}$ & 3000 & 3000 & How many epochs \\
 $\alpha_1$ & $3.06\times 10^{-4}$ & $4.34\times 10^{-3}$ & Learning rate for weight update \\ 
 $m_{1}$ & 35 & 72 & Size of minibatch in weight update \\
 $\gamma_{1}$ & 0.757 & 0.9913 & Decaying multiplier for $\alpha_1$ \\
 $s_1$ & 513 & 7 & Apply the decay multiplier for $\alpha_1$ after this many update steps \\
 %\hline
 $\alpha_2$ & $4.88\times 10^{-3}$ & - & Learning rate for architectural update \\ 
 $m_{2}$ & 142 & - & Size of minibatch in architectural update \\
 $\gamma_{2}$ & 0.701 & - & Decaying multiplier for $\alpha_2$ \\
 $s_2$ & 918 & - & Apply the decay multiplier for $\alpha_2$ after this many epochs \\
 $p_{val}$ & 2 & 1 & Going through the validation dataset this many times in one epoch \\
 
 \hline
\end{tabular}
\end{center}
\caption{List of hyperparameters used in training the emulators}
\label{tab:hparams-table}
\end{table*}

\subsection{\label{subsec:other-ml} Other emulator builder methods}

In training the emulators using non-deep learning methods, we employed the scikit-learn library \cite{pedregosa2011scikit} in Python.
We use the default parameters suggested in the library to build the emulators for all cases.

For models that can only predict a single output, an ensemble of models are trained to predict different outputs in one simulation.
For CycleGAN in the ICF JAG and ICF JAG Scalars cases, the model was trained and obtained according to ref~\cite{anirudh2019cyclegan, anirudh2020improved}.
%We believe this would give a fair comparison as our hyperparameters were chosen only by optimizing them for one case (i.e. OTS).
%We used the same hyperparameters for other case without fine-tuning them for each case.

\subsection{\label{subsec:inv-problem} Solving inverse problems with emulators}

The parameter retrieval processes for XRTS and Halo to produce Figure \ref{fig:inv-opt-results} were done using the CMA-ES \cite{hansen2016-cmaes} algorithm with population size 32 and 1200 maximum function evaluations.
Default parameters suggested in ref.\cite{hansen2016-cmaes} were used.
To give a fair results comparison, we used the same algorithm parameters and conditions in parameter retrievals via simulations and emulators.

For the Bayesian posterior sampling process, we employed the affine-invariant ensemble MCMC algorithm \cite{goodman2010-ensemble-mcmc} with 256 walkers to collect 200,000 samples for XRTS and XES cases.
The likelihood is uniform when the generated spectrum lies in a given band and it is zero when it lies outside the band.
The band is 0.035 J/keV/ster in XES and 3.5\% in XRTS as used in ref\cite{kasim-18-ipi}.
The justification of this form of likelihood is also provided in the supplementary materials of ref \cite{kasim-18-ipi}.

\subsection{Derivation of prediction uncertainty}
The randomization of the network architecture can be seen as a special case of dropout.
Therefore, the derivation of the prediction uncertainty follows the derivation in Monte Carlo (MC) dropout very closely \cite{gal2016dropout}.

Denote the input and output from the training dataset as $\mathbf{X}$ and $\mathbf{Y}$ respectively and write $\boldsymbol{\omega}$ as the active weights in the neural network.
Given the training data, $\mathbf{X}$ and $\mathbf{Y}$, the posterior distribution of the weights in the neural network can be written as
\begin{equation}
    \mathbb{P}(\boldsymbol{\omega} | \mathbf{X}, \mathbf{Y}) = \frac{\mathbb{P}(\mathbf{Y}|\mathbf{X}, \boldsymbol{\omega}) \mathbb{P}(\boldsymbol{\omega})}{Z}
\end{equation}
where $Z$ is the normalization constant, $\mathbb{P}(\mathbf{Y}|\mathbf{X}, \boldsymbol{\omega})$ is the likelihood of observing $\mathbf{Y}$ with weights $\boldsymbol{\omega}$, and $\mathbb{P}(\boldsymbol{\omega})$ is the prior distribution of the weights.

The posterior distribution of the weights are intractable, so we need to use variational inference in making the approximation to the posterior distribution.
Let the variational distribution takes the form of $\mathbb{Q}(\boldsymbol{\omega})$ where
\begin{align}
    \boldsymbol{\omega_{ij}} &= \mathbf{w_{ij}} z_{ij},\ \mathrm{and}\\
    z_{ij} &\sim \mathrm{Bernoulli}[p_{ij}(b_{ij})]
\end{align}
where $\mathbf{w_{ij}}$ is the weights of layers $j$ in group $i$, $p_{ij}$ is the probability of being selected as a function of the network variable, $b_{ij}$, as written in equation \ref{eq:prob-equation}.

In order to get the best approximation of the posterior distribution $\mathbb{P}(\boldsymbol{\omega}|\mathbf{X}, \mathbf{Y})$ with $\mathbb{Q}(\boldsymbol{\omega})$, the Kullback-Leibler (KL) divergence should be minimized.
The KL divergence to be minimized can be expressed as
\begin{equation}
\label{eq:kl-uncertainty}
    \mathrm{KL} = -\int \mathbb{Q}(\boldsymbol{\omega}) \log\left[\mathbb{P}(\mathbf{Y}|\mathbf{X},\boldsymbol{\omega})\right]\mathrm{d}\boldsymbol{\omega} + \mathrm{KL}\left[\mathbb{Q}(\boldsymbol{\omega})||\mathbb{P}(\boldsymbol{\omega})\right].
\end{equation}
The integral on the first term on the right hand side can be approximated by drawing samples from $\boldsymbol{\omega_n} \sim \mathbb{Q}(\boldsymbol{\omega})$ and performing the Monte Carlo integration on the negative log-likelihood, $-\log\left[\mathbb{P}(\mathbf{Y}|\mathbf{X},\boldsymbol{\omega_n})\right]$.
The second term on the right hand side is approximated to be $\sum_{ij} \frac{p_{ij}l}{2} ||\mathbf{w_{ij}}||^2$ where $l$ is the prior assumption of the length scale of the distribution.
We can take the prior assumption of small length scale to be able to capture high variability region better and therefore making the second term small.

With various approximation above, the KL divergence in equation \ref{eq:kl-uncertainty} to be minimized can be expressed as
\begin{equation}
\label{eq:kl-approximation-final}
    \mathrm{KL}\approx -\frac{1}{N}\sum_n \log\left[\mathbb{P}(\mathbf{Y}|\mathbf{X},\boldsymbol{\omega_n})\right].
\end{equation}
By defining the negative log likelihood as the Huber loss function, we obtain that minimizing the KL divergence in the equation \ref{eq:kl-approximation-final} is equivalent to minimizing the loss function in equations \ref{eq:weights-update} and \ref{eq:update-network-var}.
Therefore, the optimized parameters after the training can be used to approximate the posterior distribution of the weights in the form of $\mathbb{Q}(\boldsymbol{\omega})$.

%\input{additional-results.tex}

%\section*{\label{subsec:data-code-availability} Data and code availability}
%The dataset for ICF JAG and ICF JAG Scalar can be found in ref~\cite{anirudh2019cyclegan}.
%The rest of the dataset and code for emulators are available in http://doi.org/10.5281/zenodo.3782843.
%\revision{For cases with a few number of datasets (e.g. MOPS), the data is replicated to keep the number of update steps similar to other cases.}
%As we did not set a specific random seed during our research, the training results might be slightly different from those presented in this paper.

\section*{Conflict of interest}
The authors declare no conflict of interest.

\section*{Author contributions}
M.F.K. initiated the project with contributions from S.M.V.
D.W-P., L.D., S.O., P.H., D.H.F., G.G., M.J., S.K., J.K., J.T-M., and E.V. adapted and prepared the test cases.
M.F.K. designed and trained the deep neural networks and performed the analysis.
M.F.K. and S.M.V. prepared the manuscript with contributions from D.W-P., S.O., P.H., G.G., J.K., and E.V.

\section*{Acknowledgement}
M.F.K. and S.M.V. acknowledge support from the UK EPSRC grant EP/P015794/1 and the Royal Society. S.M.V. is a Royal Society University Research Fellow.
G.G. acknowledges support from AWE plc., and the UK EPSRC (EP/M022331/1 and EP/N014472/1).
E.V. is grateful for support from the European Research Council (ERC) under the European Union's Horizon 2020 research and innovation programme (grant agreement No 805162).
D.W.P. and L.D. acknowledge funding from the Natural Environment Research Council (NERC) NE/P013406/1 (A-CURE).

\bibliographystyle{unsrt}
\bibliography{bibliography}

\begin{thebibliography}{10}

\bibitem{greeley2006computational-hts}
Jeff Greeley, Thomas~F Jaramillo, Jacob Bonde, IB~Chorkendorff, and Jens~K
  N{\o}rskov.
\newblock Computational high-throughput screening of electrocatalytic materials
  for hydrogen evolution.
\newblock {\em Nature materials}, 5(11):909--913, 2006.

\bibitem{lee-2009-thomson-scattering}
HJ~Lee, P~Neumayer, J~Castor, T~D{\"o}ppner, RW~Falcone, C~Fortmann, BA~Hammel,
  AL~Kritcher, OL~Landen, RW~Lee, et~al.
\newblock X-ray thomson-scattering measurements of density and temperature in
  shock-compressed beryllium.
\newblock {\em Physical review letters}, 102(11):115001, 2009.

\bibitem{galdon2018beam-elms}
J~Galdon-Quiroga, Manuel Garcia-Munoz, KG~McClements, M~Nocente, M~Hoelzl,
  AS~Jacobsen, F~Orain, JF~Rivero-Rodriguez, Mirko Salewski, L~Sanchis-Sanchez,
  et~al.
\newblock Beam-ion acceleration during edge localized modes in the asdex
  upgrade tokamak.
\newblock {\em Physical review letters}, 121(2):025002, 2018.

\bibitem{peterson2017zonal}
JL~Peterson, KD~Humbird, JE~Field, ST~Brandon, SH~Langer, RC~Nora, BK~Spears,
  and PT~Springer.
\newblock Zonal flow generation in inertial confinement fusion implosions.
\newblock {\em Physics of Plasmas}, 24(3):032702, 2017.

\bibitem{kwan2015cosmic-emu}
Juliana Kwan, Katrin Heitmann, Salman Habib, Nikhil Padmanabhan, Earl Lawrence,
  Hal Finkel, Nicholas Frontiere, and Adrian Pope.
\newblock Cosmic emulation: fast predictions for the galaxy power spectrum.
\newblock {\em The Astrophysical Journal}, 810(1):35, 2015.

\bibitem{brockherde2017bypassing}
Felix Brockherde, Leslie Vogt, Li~Li, Mark~E Tuckerman, Kieron Burke, and
  Klaus-Robert M{\"u}ller.
\newblock Bypassing the kohn-sham equations with machine learning.
\newblock {\em Nature communications}, 8(1):872, 2017.

\bibitem{rupp2012fast-molecular-energy-krr}
Matthias Rupp, Alexandre Tkatchenko, Klaus-Robert M{\"u}ller, and O~Anatole
  Von~Lilienfeld.
\newblock Fast and accurate modeling of molecular atomization energies with
  machine learning.
\newblock {\em Physical review letters}, 108(5):058301, 2012.

\bibitem{ulyanov2018deep-image-prior}
Dmitry Ulyanov, Andrea Vedaldi, and Victor Lempitsky.
\newblock Deep image prior.
\newblock In {\em Proceedings of the IEEE Conference on Computer Vision and
  Pattern Recognition}, pages 9446--9454, 2018.

\bibitem{pham2018efficient-enas}
Hieu Pham, Melody~Y Guan, Barret Zoph, Quoc~V Le, and Jeff Dean.
\newblock Efficient neural architecture search via parameter sharing.
\newblock {\em arXiv preprint arXiv:1802.03268}, 2018.

\bibitem{cai2018proxylessnas}
Han Cai, Ligeng Zhu, and Song Han.
\newblock Proxylessnas: Direct neural architecture search on target task and
  hardware.
\newblock {\em arXiv preprint arXiv:1812.00332}, 2018.

\bibitem{he2016deep-resnet}
Kaiming He, Xiangyu Zhang, Shaoqing Ren, and Jian Sun.
\newblock Deep residual learning for image recognition.
\newblock In {\em Proceedings of the IEEE conference on computer vision and
  pattern recognition}, pages 770--778, 2016.

\bibitem{huber1992robust-loss}
Peter~J Huber.
\newblock Robust estimation of a location parameter.
\newblock In {\em Breakthroughs in statistics}, pages 492--518. Springer, 1992.

\bibitem{williams1992simple-REINFORCE}
Ronald~J Williams.
\newblock Simple statistical gradient-following algorithms for connectionist
  reinforcement learning.
\newblock {\em Machine learning}, 8(3-4):229--256, 1992.

\bibitem{hansen2016-cmaes}
Nikolaus Hansen.
\newblock The cma evolution strategy: A tutorial.
\newblock {\em arXiv preprint arXiv:1604.00772}, 2016.

\bibitem{gregori2003theoretical-thomson-scattering}
G~Gregori, Siegfried~H Glenzer, W~Rozmus, RW~Lee, and OL~Landen.
\newblock Theoretical model of x-ray scattering as a dense matter probe.
\newblock {\em Physical Review E}, 67(2):026412, 2003.

\bibitem{tzeferacos2018laboratory-ots}
P~Tzeferacos, A~Rigby, AFA Bott, AR~Bell, R~Bingham, A~Casner, F~Cattaneo,
  EM~Churazov, J~Emig, F~Fiuza, et~al.
\newblock Laboratory evidence of dynamo amplification of magnetic fields in a
  turbulent plasma.
\newblock {\em Nature communications}, 9(1):591, 2018.

\bibitem{regan2013hot-xes}
SP~Regan, R~Epstein, BA~Hammel, LJ~Suter, HA~Scott, MA~Barrios, DK~Bradley,
  DA~Callahan, C~Cerjan, GW~Collins, et~al.
\newblock Hot-spot mix in ignition-scale inertial confinement fusion targets.
\newblock {\em Physical review letters}, 111(4):045001, 2013.

\bibitem{ciricosta2017simultaneous-xes}
Orlando Ciricosta, H~Scott, P~Durey, BA~Hammel, R~Epstein, TR~Preston,
  SP~Regan, SM~Vinko, NC~Woolsey, and JS~Wark.
\newblock Simultaneous diagnosis of radial profiles and mix in nif
  ignition-scale implosions via x-ray spectroscopy.
\newblock {\em Physics of Plasmas}, 24(11):112703, 2017.

\bibitem{hatfield2016galaxy-halo}
PW~Hatfield, SN~Lindsay, MJ~Jarvis, B~H{\"a}u{\ss}ler, M~Vaccari, and A~Verma.
\newblock The galaxy--halo connection in the video survey at 0.5< z< 1.7.
\newblock {\em Monthly Notices of the Royal Astronomical Society},
  459(3):2618--2631, 2016.

\bibitem{korenaga2012seismic-mcmc}
J~Korenaga and WW~Sager.
\newblock Seismic tomography of shatsky rise by adaptive importance sampling.
\newblock {\em Journal of Geophysical Research: Solid Earth}, 117(B8), 2012.

\bibitem{tegen2019global-climate-model-aerosol}
Ina Tegen, David Neubauer, Sylvaine Ferrachat, Siegenthaler-Le Drian, Isabelle
  Bey, Nick Schutgens, Philip Stier, Duncan Watson-Parris, Tanja Stanelle,
  Hauke Schmidt, et~al.
\newblock The global aerosol-climate model echam6. 3-ham2. 3-part 1: Aerosol
  evaluation.
\newblock {\em Geoscientific Model Development}, 12(4):1643--1677, 2019.

\bibitem{khatiwala2007computational}
Samar Khatiwala.
\newblock A computational framework for simulation of biogeochemical tracers in
  the ocean.
\newblock {\em Global Biogeochemical Cycles}, 21(3), 2007.

\bibitem{anirudh2019cyclegan}
Rushil Anirudh, Peer-Timo Bremer, Jayaraman~Jayaraman Thiagrarjan, and USDOE
  National Nuclear~Security Administration.
\newblock Cycle consistent surrogate for inertial confinement fusion.
\newblock 2 2019.

\bibitem{loshchilov2016-cmaes-hyperparameters-tuning}
Ilya Loshchilov and Frank Hutter.
\newblock Cma-es for hyperparameter optimization of deep neural networks.
\newblock {\em arXiv preprint arXiv:1604.07269}, 2016.

\bibitem{pedregosa2011scikit}
Fabian Pedregosa, Ga{\"e}l Varoquaux, Alexandre Gramfort, Vincent Michel,
  Bertrand Thirion, Olivier Grisel, Mathieu Blondel, Peter Prettenhofer, Ron
  Weiss, Vincent Dubourg, et~al.
\newblock Scikit-learn: Machine learning in python.
\newblock {\em Journal of machine learning research}, 12(Oct):2825--2830, 2011.

\bibitem{anirudh2020improved}
Rushil Anirudh, Jayaraman~J Thiagarajan, Peer-Timo Bremer, and Brian~K Spears.
\newblock Improved surrogates in inertial confinement fusion with manifold and
  cycle consistencies.
\newblock {\em Proceedings of the National Academy of Sciences}, 2020.

\bibitem{kasim-18-ipi}
MF~Kasim, TP~Galligan, J~Topp-Mugglestone, G~Gregori, and SM~Vinko.
\newblock Inverse problem instabilities in large-scale modeling of matter in
  extreme conditions.
\newblock {\em Physics of Plasmas}, 26(11):112706, 2019.

\bibitem{wierstra2008-natural-nes}
Daan Wierstra, Tom Schaul, Jan Peters, and Juergen Schmidhuber.
\newblock Natural evolution strategies.
\newblock In {\em 2008 IEEE Congress on Evolutionary Computation (IEEE World
  Congress on Computational Intelligence)}, pages 3381--3387. IEEE, 2008.

\bibitem{goodman2010-ensemble-mcmc}
Jonathan Goodman and Jonathan Weare.
\newblock Ensemble samplers with affine invariance.
\newblock {\em Communications in applied mathematics and computational
  science}, 5(1):65--80, 2010.

\bibitem{srivastava2014dropout}
Nitish Srivastava, Geoffrey Hinton, Alex Krizhevsky, Ilya Sutskever, and Ruslan
  Salakhutdinov.
\newblock Dropout: a simple way to prevent neural networks from overfitting.
\newblock {\em The journal of machine learning research}, 15(1):1929--1958,
  2014.

\bibitem{gal2016dropout}
Yarin Gal and Zoubin Ghahramani.
\newblock Dropout as a bayesian approximation: Representing model uncertainty
  in deep learning.
\newblock In {\em international conference on machine learning}, pages
  1050--1059, 2016.

\bibitem{gal2017concrete}
Yarin Gal, Jiri Hron, and Alex Kendall.
\newblock Concrete dropout.
\newblock In {\em Advances in neural information processing systems}, pages
  3581--3590, 2017.

\bibitem{mcinnes2018umap}
Leland McInnes, John Healy, and James Melville.
\newblock Umap: Uniform manifold approximation and projection for dimension
  reduction.
\newblock {\em arXiv preprint arXiv:1802.03426}, 2018.

\bibitem{ronen2019convergence}
Basri Ronen, David Jacobs, Yoni Kasten, and Shira Kritchman.
\newblock The convergence rate of neural networks for learned functions of
  different frequencies.
\newblock In {\em Advances in Neural Information Processing Systems}, pages
  4763--4772, 2019.

\bibitem{chowdhury2019efficient}
Drimik~Roy Chowdhury and Muhammad~Firmansyah Kasim.
\newblock Efficient parameter sampling for neural network construction.
\newblock {\em arXiv preprint arXiv:1912.10559}, 2019.

\bibitem{kates2019predicting-tokamak}
Julian Kates-Harbeck, Alexey Svyatkovskiy, and William Tang.
\newblock Predicting disruptive instabilities in controlled fusion plasmas
  through deep learning.
\newblock {\em Nature}, 568(7753):526--531, 2019.

\bibitem{emma2018machine-learning-particle-accelerator}
C~Emma, A~Edelen, MJ~Hogan, B~O’Shea, G~White, and V~Yakimenko.
\newblock Machine learning-based longitudinal phase space prediction of
  particle accelerators.
\newblock {\em Physical Review Accelerators and Beams}, 21(11):112802, 2018.

\bibitem{liu2013gp-surrogate-optimization}
Bo~Liu, Qingfu Zhang, and Georges~GE Gielen.
\newblock A gaussian process surrogate model assisted evolutionary algorithm
  for medium scale expensive optimization problems.
\newblock {\em IEEE Transactions on Evolutionary Computation}, 18(2):180--192,
  2013.

\bibitem{shahriari2015bayesopt}
Bobak Shahriari, Kevin Swersky, Ziyu Wang, Ryan~P Adams, and Nando De~Freitas.
\newblock Taking the human out of the loop: A review of bayesian optimization.
\newblock {\em Proceedings of the IEEE}, 104(1):148--175, 2015.

\bibitem{kritcher-2008-thomson-scattering}
Andrea~L Kritcher, Paul Neumayer, John Castor, Tilo D{\"o}ppner, Roger~W
  Falcone, Otto~L Landen, Hae~Ja Lee, Richard~W Lee, Edward~C Morse, Andrew Ng,
  et~al.
\newblock Ultrafast x-ray thomson scattering of shock-compressed matter.
\newblock {\em Science}, 322(5898):69--71, 2008.

\bibitem{zohm1996edge-elms}
Hartmut Zohm.
\newblock Edge localized modes (elms).
\newblock {\em Plasma Physics and Controlled Fusion}, 38(2):105, 1996.

\bibitem{cavedon2017pedestal-elms}
M~Cavedon, T~P{\"u}tterich, Eleonora Viezzer, FM~Laggner, A~Burckhart, M~Dunne,
  R~Fischer, A~Lebschy, F~Mink, U~Stroth, et~al.
\newblock Pedestal and e r profile evolution during an edge localized mode
  cycle at asdex upgrade.
\newblock {\em Plasma Physics and Controlled Fusion}, 59(10):105007, 2017.

\bibitem{viezzer2018ion-elms-astra}
E~Viezzer, M~Cavedon, E~Fable, FM~Laggner, RM~McDermott, J~Galdon-Quiroga,
  MG~Dunne, A~Kappatou, C~Angioni, P~Cano-Megias, et~al.
\newblock Ion heat transport dynamics during edge localized mode cycles at
  asdex upgrade.
\newblock {\em Nuclear Fusion}, 58(2):026031, 2018.

\bibitem{fable2013novel-elms-astra}
E~Fable, C~Angioni, FJ~Casson, D~Told, AA~Ivanov, F~Jenko, RM~McDermott, S~Yu
  Medvedev, GV~Pereverzev, F~Ryter, et~al.
\newblock Novel free-boundary equilibrium and transport solver with
  theory-based models and its validation against asdex upgrade current ramp
  scenarios.
\newblock {\em Plasma Physics and Controlled Fusion}, 55(12):124028, 2013.

\bibitem{willensdorfer2013particle-elms}
M~Willensdorfer, E~Fable, E~Wolfrum, Leena Aho-Mantila, F~Aumayr, R~Fischer,
  F~Reimold, F~Ryter, et~al.
\newblock Particle transport analysis of the density build-up after the l--h
  transition in asdex upgrade.
\newblock {\em Nuclear Fusion}, 53(9):093020, 2013.

\bibitem{halomodsoftware}
Steven Murray.
\newblock {halomod: Python package for dealing with the Halo Model}, June 2017.

\bibitem{wake2011galaxy}
David~A Wake, Katherine~E Whitaker, Ivo Labb{\'e}, Pieter~G Van~Dokkum, Marijn
  Franx, Ryan Quadri, Gabriel Brammer, Mariska Kriek, Britt~F Lundgren, Danilo
  Marchesini, et~al.
\newblock Galaxy clustering in the newfirm medium band survey: the relationship
  between stellar mass and dark matter halo mass at 1< z< 2.
\newblock {\em The Astrophysical Journal}, 728(1):46, 2011.

\bibitem{kriest2015mops}
Iris Kriest and Andreas Oschlies.
\newblock Mops-1.0: modelling the regulation of the global oceanic nitrogen
  budget by marine biogeochemical processes.
\newblock {\em Geoscientific Model Development}, 8:2929--2957, 2015.

\bibitem{samarkhatiwala_2018_computational}
samarkhatiwala.
\newblock samarkhatiwala/tmm: Version 2.0 of the transport matrix method
  software, May 2018.

\bibitem{marshall1997finite-mitgcm}
John Marshall, Alistair Adcroft, Chris Hill, Lev Perelman, and Curt Heisey.
\newblock A finite-volume, incompressible navier stokes model for studies of
  the ocean on parallel computers.
\newblock {\em Journal of Geophysical Research: Oceans}, 102(C3):5753--5766,
  1997.

\bibitem{kriest2017calibrating-mops}
Iris Kriest, Volkmar Sauerland, Samar Khatiwala, Anand Srivastav, and Andreas
  Oschlies.
\newblock Calibrating a global three-dimensional biogeochemical ocean model
  (mops-1.0).
\newblock {\em Geoscientific Model Development}, 10:127--154, 2017.

\end{thebibliography}

\end{document}